%% file: pscc2024_paper.tex
\documentclass{IEEEtran4PSCC}
\ifCLASSINFOpdf
   \usepackage[pdftex]{graphicx}
\else
   \usepackage[dvips]{graphicx}
\fi
\usepackage{amsfonts,amsmath,amssymb,amsthm}
\usepackage{cases}
\usepackage{cleveref}
\usepackage{multirow}
\usepackage[table,xcdraw]{xcolor}
\usepackage{booktabs}
\usepackage{float}
\floatstyle{ruled}
\newfloat{model}{thp}{lop}
\floatname{model}{Model}

\usepackage{todonotes}

\input{tex/definitions}

\makeatletter
\let\old@ps@headings\ps@headings
\let\old@ps@IEEEtitlepagestyle\ps@IEEEtitlepagestyle
\def\psccfooter#1{%
    \def\ps@headings{%
        \old@ps@headings%
        \def\@oddfoot{\strut\hfill#1\hfill\strut}%
        \def\@evenfoot{\strut\hfill#1\hfill\strut}%
    }%
    \def\ps@IEEEtitlepagestyle{%
        \old@ps@IEEEtitlepagestyle%
        \def\@oddfoot{\strut\hfill#1\hfill\strut}%
        \def\@evenfoot{\strut\hfill#1\hfill\strut}%
    }%
    \ps@headings%
}
\makeatother

\psccfooter{%
        \parbox{\textwidth}{\hrulefill \\ \small{22nd Power Systems Computation Conference} \hfill \begin{minipage}{0.2\textwidth}\centering \vspace*{4pt} \includegraphics[scale=0.06]{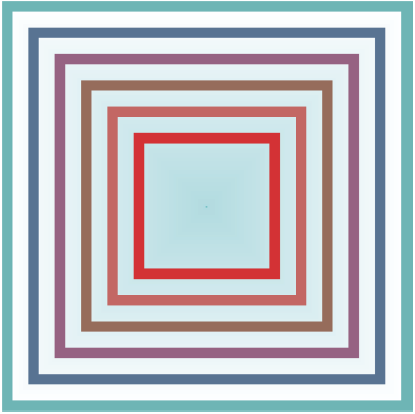}\\\small{PSCC 2024} \end{minipage} \hfill \small{Paris, France --- June 4 -- June 7, 2024}}%
}

\begin{document}
%
\title{Scalable Exact Verification of Optimization Proxies for Large-Scale Optimal Power Flow}

\author{
\IEEEauthorblockN{Rahul Nellikkath \\ Spyros Chatzivasileiadis}
\IEEEauthorblockA{Denmark Technical University\\
\{rnelli,spchatz\}@dtu.dk}
\and
\IEEEauthorblockN{Mathieu Tanneau\\ Pascal Van Hentenryck}
\IEEEauthorblockA{Georgia Institute of Technology \\  \{mathieu.tanneau, pascal.vanhentenryck\}@isye.gatech.edu }
}

\maketitle

\begin{abstract}
    Optimal Power Flow (OPF) is a valuable tool for power system operators, but it is a difficult problem to solve for large systems.
    Machine Learning (ML) algorithms, especially Neural Networks-based (NN) optimization proxies, have emerged as a promising new tool for solving OPF, by estimating the OPF solution much faster than traditional methods.
    However, these ML algorithms act as black boxes, and it is hard to assess their worst-case performance across the entire range of possible inputs than an OPF can have.
    Previous work has proposed a mixed-integer programming-based methodology to quantify the worst-case violations caused by a NN trained to estimate the OPF solution, throughout the entire input domain.
    This approach, however, does not scale well to large power systems and more complex NN models.
    This paper addresses these issues by proposing a scalable algorithm to compute worst-case violations of NN proxies used for approximating large power systems within a reasonable time limit.
    This will help build trust in ML models to be deployed in large industry-scale power grids.
\end{abstract}


\thanksto{\noindent This research was partly supported by NSF award 2112533 and by the ERC Starting Grant VeriPhIED, funded by the European Research Council, Grant Agreement 949899.}

\input{tex/introduction}

\input{tex/verification}

\input{tex/accelarating_verification}




\input{tex/results}

\section{Conclusion}
This paper proposes two methods to accelerate the exact verification of the NNs for Large-Scale Optimal Power Flow. Firstly, we propose a gradient-based method for finding adversarial test cases for OPF proxies that can find test cases that are close to the worst-case constraint violations. Secondly, we show the benefits of using advanced bound-tightening techniques, in particular $\ABC$, to tighten the MIP dual bounds. The proposed methodology converges on the \ieee{} and \pegaseS{} systems, which was impossible with the state-of-the-art OBBT techniques. This is significant, as it makes it possible to use formal verification to ensure the safety and reliability of power grids. In the future, more effective cutting planes will be explored to narrow the gap between primal and dual bounds.

\bibliographystyle{IEEEtran}
\bibliography{Referance}

\end{document}

%% file: tex/definitions.tex
\newcommand{\ABC}{\alpha,\beta{-}\text{CROWN}}

\newcommand{\e}{\mathbf{e}}  
\newcommand{\pd}{\mathbf{p}^{\text{d}}}
\newcommand{\pf}{\mathbf{p}^{\text{f}}}
\newcommand{\pfmax}{\mathbf{\overline{p}}^{\text{f}}}
\newcommand{\pg}{\mathbf{p}^{\text{g}}}
\newcommand{\pgmin}{\mathbf{\underline{p}}^{\text{g}}}
\newcommand{\pgmax}{\mathbf{\overline{p}}^{\text{g}}}
\newcommand{\PTDF}{\Phi}
\newcommand{\pdmin}{\mathbf{\underline{p}}^{\text{d}}}
\newcommand{\pdmax}{\mathbf{\overline{p}}^{\text{d}}}

\newcommand{\pghat}{\mathbf{\hat{p}}^{\text{g}}}
\newcommand{\pdadv}{\mathbf{\tilde{p}}^{\text{d}}}  

%% file: tex/introduction.tex
\section{Introduction}

The Optimal Power Flow (OPF) problem \cite{CARPENTIER1979} is a
fundamental problem in power systems operations. It is routinely used
for clearing day-ahead and real-time electricity markets, for
planning, to compute optimal control setpoints, and to conduct
reliability analyses.  However, in its original form, the AC-OPF
problem is often challenging to solve as it is a non-linear and
non-convex optimization.  Recent years have witnessed a surge of
interest in optimization proxies for OPF, i.e., Machine Learning (ML)
models that approximate the input/output mapping of OPF \cite{MLDCOPF}
\cite{ReviewMLSec,SurveyML,recent}.  Once trained, optimization
proxies can produce high-quality, close-to-feasible solutions to OPF
in milliseconds, thereby overcoming the computational barrier of
AC-OPF for large, industry-scale power grids.  Additionally, trained
ML models have shown great potential in serving as fast surrogate
functions to replace computationally intensive constraints
\cite{surrogate} or bi-level optimization problems.

Nevertheless, the black-box nature of optimization proxies poses a
significant challenge to their deployment and adoption in industry.
In general, optimization proxies come without formal guarantees across
the entire range of possible inputs, which is particularly concerning
considering the potential use of OPF in safety-critical applications.
This has motivated the use of Neural Network (NN) verification to
quantify an optimization proxy's worst-case performance \cite{Andreas}.

State-of-the-art approaches for NN verification first build a
mixed-integer programming (MIP) representation of the NN, then solve a
MIP problem to provably compute its worst-case performance across a
given input domain \cite{Andreas,MYDC,MYAC}.  Prior works have
demonstrated how various worst-case guarantees can be computed for
DC-OPF \cite{Andreas,MYDC} and AC-OPF problems\cite{MYAC,sam_stt}.
Furthermore, prior research has demonstrated how to use NN
verification and worst-case guarantees to build a trustworthy NN
\cite{MinWC,Enrich}.

While existing verification approaches leverage mature MIP technology,
their high computational cost has so far limited their applications to
small power grids and simple NN models. This research is a step in
addressing this limitation. It proposes new primal and dual
acceleration techniques to improve the scalability of OPF proxies
verification and demonstrates their performance on large power
grids. More precisely, the paper makes three contributions:
\begin{enumerate}
\item The paper shows how to find adversarial test cases 
  quickly, by using gradient-based strategies that exploit the
  structure of OPF problems. These test cases can then be used to
  warm-start the verification MIP and obtain better primal bounds
  faster.

\item The paper shows the benefits of advanced bound-tightening
  techniques and, in particular, $\ABC$
  \cite{ABC1,ABC2,betaCrown,chevalier2023gpuaccelerated} to tighten
  the MIP dual bounds. Again, these techniques leverage the properties
  of OPF problems to reduce the complexity of the MIP problem and help
  the dual bound converge faster.

\item The paper shows that these acceleration techniques makes it
  possible to formally verify OPF proxies for large power grids that
  are an order of magnitude larger than those used in existing approaches.  
\end{enumerate}

This rest of the paper is structured as follows. Section II describes
the DC-OPF \cite{DCOPF} problem and the optimization algorithm used to
determine the worst-case guarantees. Section III introduces the
proposed methods for accelerating the NN verification. Section IV
provides the simulation setup used and reports the results
demonstrating the performance of proposed techniques. Section V
discusses the possible opportunities to further improve the system performance
and draw the conclusions of the paper.

%% file: tex/verification.tex
\section{DNN Verification for OPF Proxies}
\label{sec:verification}

The paper proposes new acceleration techniques to address existing
scalability challenges in verifying OPF proxies, both on the primal
and dual sides.  The methodology is evaluated on DC-OPF proxies but,
ultimately, similar approaches can be applied to verify AC-OPF
proxies.  In the following, the set of buses is denoted by $\mathcal{N}
\, {=} \, \{1, ..., N\}$ and the set of branches by $\mathcal{E} \,
   {=} \, \{1, ..., E\}$.  The vector of all ones is denoted by $\e \,
   {=} \, (1, ..., 1)$.  For ease of reading and without loss of
   generality, the presentation assumes that one generator and one
   load are attached to each bus, and that the costs are linear.
   Unless specified otherwise, all quantities are in per-unit (p.u.).

\subsection{DC Optimal Power Flow}
\label{sec:verification:OPF}

The DC optimal power flow (DC-OPF) is a linear approximation of
AC-OPF, where voltage magnitudes are assumed to be 1p.u. and voltage
angles are assumed to be small, losses are neglected, reactive power
is ignored, and the power flow equations are linearized.  The DC-OPF
model underlies most electricity markets in the US.

Model \ref{model:DC-OPF} presents the DC-OPF linear programming (LP)
model.  The model takes, as inputs, the cost $c$, the demand $\pd$,
the lower and upper generation limits $\pgmin, \pgmax$, the Power
Transfer Distribution Factor (PTDF) matrix $\PTDF \, {\in} \,
\mathbb{R}^{E \times N}$, and the thermal limits $\pfmax$.  Its
variables are the generation dispatches, which are denoted by $\pg$.
The objective \eqref{eq:DCOPF:Obj} minimizes the cost of generation.
Constraint \eqref{eq:DCOPF:con:power_balance} enforces the power balance
in the system, i.e., total generation equals total demand.
Constraints \eqref{eq:DCOPF:con:generation_limits} enforce minimum and
maximum limits on generation dispatch.  Constraints
\eqref{eq:DCOPF:con:thermal_constraints} enforce thermal constraints
on each branch.

\begin{model}[!t]
    \caption{The DC-OPF Model}
    \label{model:DC-OPF}
    \begin{subequations}
    \label{eq:DCOPF}
    \begin{align}
        \min \quad & c^{\top} \pg \label{eq:DCOPF:Obj}\\
        \text{s.t.} \quad
            & \e^{\top} \pg = \e^{\top} \pd
                \label{eq:DCOPF:con:power_balance}\\
            & \pgmin \leq \pg \leq \pgmax
                \label{eq:DCOPF:con:generation_limits} \\
            & | \PTDF (\pg - \pd) | \leq \pfmax 
                \label{eq:DCOPF:con:thermal_constraints}
    \end{align}
    \end{subequations}
\end{model}



\subsection{Neural Networks for Optimal Power Flow Approximation}
\label{sec:verification:DNN}

\begin{figure}[!t]
\centerline{\includegraphics[scale=.34]{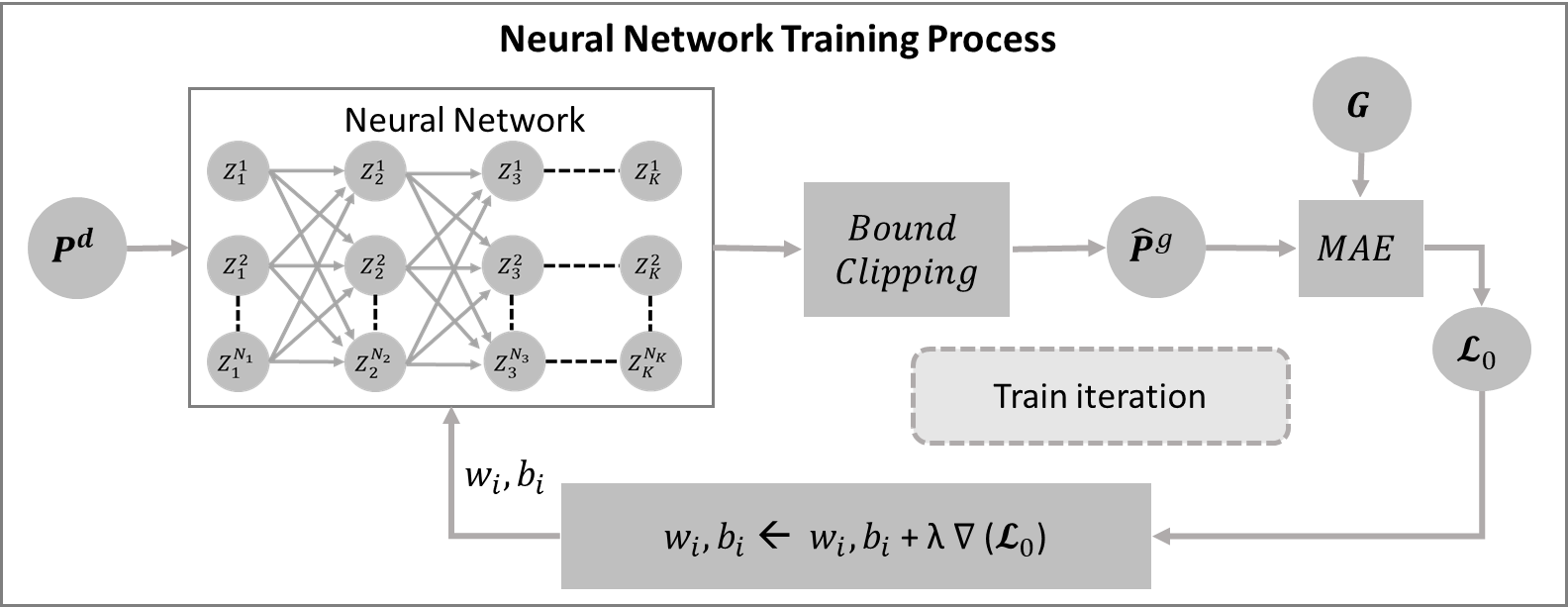}}
\caption{Illustration of the neural network architecture to predict the optimal generation active power outputs $\pghat$ using the active power demand $\pd$ as input: There are K hidden layers in the neural network with $N_k$ neurons each. Where k = 1, ...,K.}
\label{NN_basic}
\end{figure}

Neural networks (NNs) are universal function approximators.  Given
sufficient size and appropriate training, they can approximate any
function, including the mapping between power system demand and the
DC-OPF generation setpoints. 
This makes them a popular architecture for optimization proxies.
NNs are comprised of a set of
interconnected hidden layers with multiple neurons in each layer. Each
neuron in a hidden layer receives inputs from the neurons in the
previous layer and applies a linear transformation to them. The output
of this transformation is then passed through a nonlinear activation
function. The activation function introduces nonlinearity into the
network, which allows it to approximate a wider range of functions.
The final layer of the NN outputs the predicted DC-OPF generation
setpoints. The NN is trained by minimizing the error between the
predicted and actual generation setpoints. For a NN with $K$ hidden 
layers and $N_k$ number of neurons in hidden layer $k$, as
shown in \cref{NN_basic}, the information arriving at layer $k$ can be
formulated as follows:
\begin{equation}
   \hat{\mathbf{Z}}_k = \mathbf{w_{k}}\mathbf{Z}_{k-1}+\mathbf{b_{k}}\label{NN1}
\end{equation}
where ${\mathbf{Z}}_{k-1}$ is the output of the neurons in layer
$k{-}1$, $\hat{\mathbf{Z}}_k$ is the information received at layer $k$,
$\mathbf{w_{k}}$ and $\mathbf{b_{k}}$ are the weights and biases
connecting layer $k{-}1$ and $k$.

Each neuron in a NN applies a nonlinear activation function, e.g.,
the sigmoid function. The output of each layer in the NN can be
represented as follows:
\begin{equation}
    \mathbf{Z}_k = \sigma( \hat{\mathbf{Z}}_k ) \label{eq:sigma}
\end{equation}
where $\sigma$ is the nonlinear activation function. This paper uses
the rectified linear unit (ReLU) as the activation function: it can
be formulated as follows:
\begin{align}
    \mathbf{Z}_k &= \max( \hat{\mathbf{Z}}_k,0)\label{Relu}
\end{align}
The ReLU is a simple but effective activation function that has been
shown to accelerate the NN training. Furthermore, since ReLU is
piecewise linear, it can be used to estimate worst-case guarantees and
achieve exact verification.

Since active power generation is limited by generation capacities
\eqref{eq:DCOPF:con:generation_limits}, the output of the NN is clipped. The bound clipping
process can be denoted as follows:
\begin{align}
    \pghat &= \min( \max( \hat{\mathbf{Z}}_K, \mathbf{\pgmin}),\mathbf{\pgmax}) \label{BndClp}
\end{align}
This can also be implemented using two ReLUs as follows:
\begin{align}
   \pghat =  (\mathbf{\pgmax} - \sigma(\mathbf{\pgmax}- (\mathbf{\pgmin} + \sigma (\hat{\mathbf{Z}}_K-\mathbf{\pgmin}))))  \label{BndClp2}
\end{align}

Given a set of active power demands of a system, this paper trains NNs
to determine the optimal active power setpoints of the DC-OPF problem,
as a guiding application. The average error in predicting the optimal
generation setpoints in the training data set, denoted by
$\mathcal{L}_{0}$, is measured as
\begin{equation}
    \mathcal{L}_{0} = \frac{1}{N} \sum_{i=1}^N | \pg_{i} - \pghat_{i}|
    \label{L_0}
\end{equation}
where $N$ is the number of training data points,
and $\pg_{i}, \pghat_{i}$ are the ground truth (produced by an optimization solver) and predicted (by the DNN) generation dispatches, respectively.

The backpropagation algorithm modifies the weights $\mathbf w$ and
biases $\mathbf b$ in every iteration of the neural network (NN)
training to minimize the mean prediction error $\mathcal{L}_{0}$ of
the generation setpoints in the training set. However, minimizing the
mean error does not guarantee constraint compliance for all the input
combinations in the input domain. Thus, verifying the NN performance
and quantifying the worst-case guarantees is essential for deployment.

\subsection{Worst-Case Guarantees for Neural Networks}

The goal of the exact verification is to provide worst-case guarantees on
the predictions of a NN. These worst-case guarantees provide an upper
bound on constraint violations, which can be used to ensure that the
NN will always produce ``reasonable'' setpoints, even under
adversarial conditions.

To determine these worst-case guarantees, the trained NN is
reformulated into a mixed-integer linear programming (MILP) problem
using the method proposed in \cite{Andreas}. This involves converting
the non-linear activation functions into linear constraints.  The NN
formulation given in Equation (\ref{NN1}) is linear and can thus be
used directly in the MILP formulation for verification. However, the
non-linear ReLU activation (\ref{Relu}) and the bound clipping most be
reformulated to be integrated into the MILP-based verifier: 
\begin{subnumcases}
{\mathbf{Z}_k = \max(\hat{\mathbf{Z}}_k,0)\Rightarrow}
\mathbf{Z}_k  \leq \hat{\mathbf{Z}}_k - \underline{\mathbf{Z}}_k  (1-\mathbf{y}_k) \label{Relu1} \\ 
\mathbf{Z}_k  \geq \hat{\mathbf{Z}}_k \label{Relu2}   \\
\mathbf{Z}_k  \leq \overline{\mathbf{Z}}_k \mathbf{y}_k  \label{Relu3}  \\
\mathbf{Z}_k   \geq \mathbf{0}  \label{Relu4}  \\
\mathbf{y}_k \in \{0,1\}^{N_k} \label{Relu5}
\end{subnumcases}
where $\mathbf{Z}_k$ and $\hat{\mathbf{Z}}_k$ are the outputs and
inputs of the ReLU activation function, $ \underline{\mathbf{Z}}_k$
and $\overline{\mathbf{Z}}_k$ are large big-M values whose bound
constraints cannot be binding, and $\mathbf{y}_k$ is the binary
variable.

\subsubsection{Worst-Case Guarantees for Constraint Violations}
\label{sec:verification:formulation}

This section describes the MILP-based NN verifiers used to determine the
maximum line flow constraint violations and the maximum power-flow
imbalance as a result of the NN. The worst-case line flow constraint
violation in line $e$ can be determined by the MILP-based NN verifier:
\begin{subequations}
    \label{WCeq}
    \begin{align}
        v^{l}_{e} = \quad \max_{\pd} \quad 
            & \max (0, |\pf_{e}| - \pfmax_{e})\\
        \text{s.t.} \quad
            & \eqref{NN1}, \eqref{Relu1}- \eqref{Relu5} \\
            & \pf = \Phi(\pghat - \pd)\\
            & \pdmin \leq \pd \leq \pdmax
    \end{align}
\end{subequations}
The worst-case violation across all lines is then $v^{l} \, {=} \, \max_{e \in \mathcal{E}}(v^{l}_{e})$.
\noindent
The worst-case power balance violation can be formulated similarly:
\begin{subequations}
\begin{align}
    v^{\text{PB}} = \quad \max_{\pd} \quad 
        & | \e^{\top} \pd - \e^{\top}\pghat|\\
    \text{s.t.} \quad
        & \eqref{NN1}, \eqref{Relu1}- \eqref{Relu5} \\
        & \pdmin \leq \pd \leq \pdmax
\end{align}
\end{subequations}

Solving these MILP-based NN verifiers to optimality ensures that the
$v^{l}$ and $v^{\text{PB}}$ values so obtained are global optima. As a result,
they guarantee that there is no input $\pd$ in the entire
input domain leading to constraint violations larger than the obtained values.

%% file: tex/accelarating_verification.tex
\section{Acceleration Techniques}
\label{sec:acceleration}

\begin{figure}[!t]
    \centerline{\includegraphics[scale=.39]{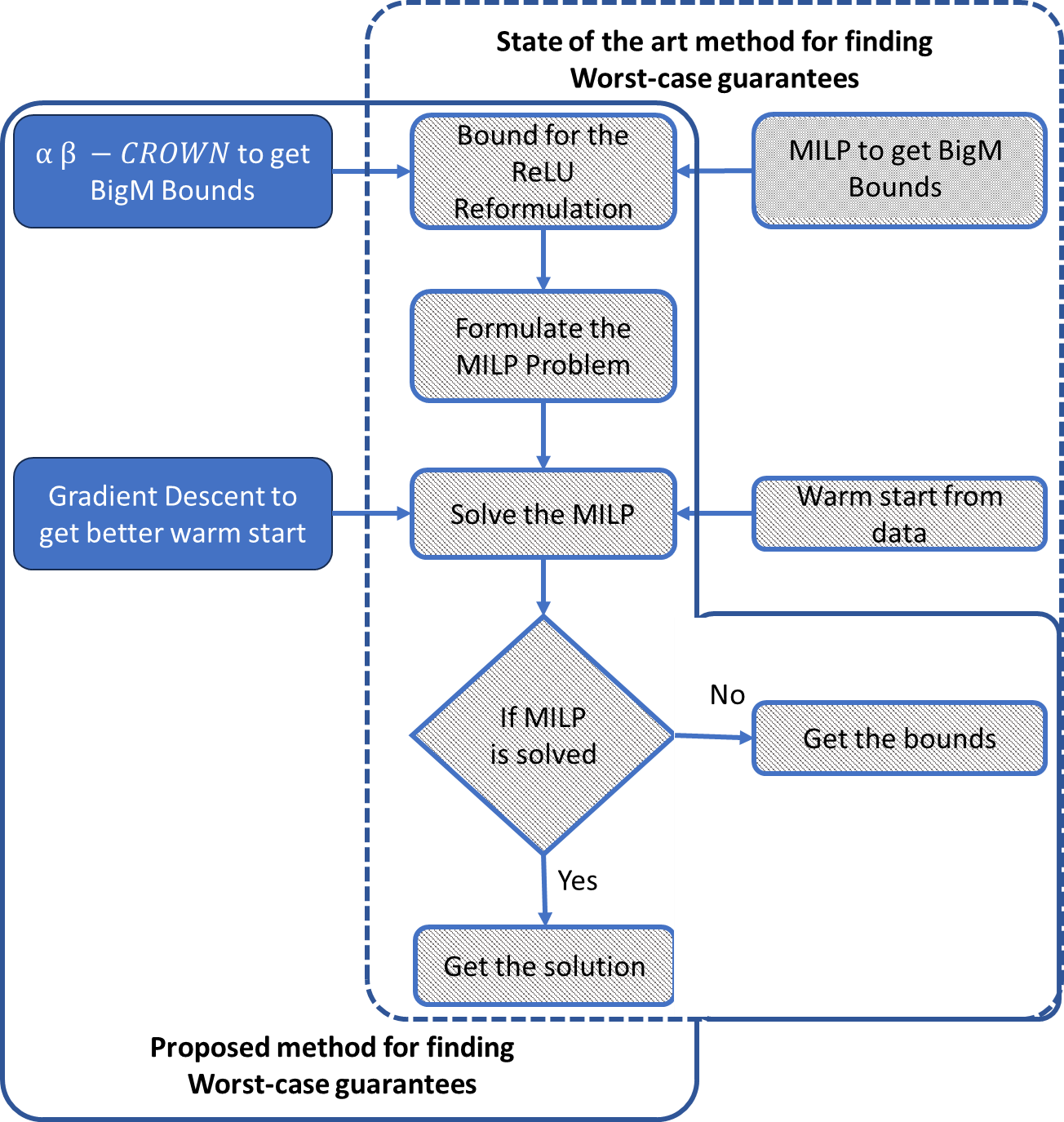}}
    \caption{Proposed Method for accelerating the NN exact verification for computing worst-case guarantees}
    \label{opt_alg}
\end{figure}

The strength of the MIP formulation directly impacts the quality of
the worst-case guarantees.  Weak MIP formulations require significant
--and costly-- branching to close the gap between primal and dual
bounds.  This paper proposes two methods for accelerating the
solving of MILP verifiers that computes worst-case guarantees for NNs.
The overall methodology is presented in \cref{opt_alg}, which
also contrasts the proposed and prior approaches.

There are two innovations in the proposed approaches, addressing dual
and primal bounds of the MILP-based NN verifiers respectively.  The
first innovation is the use of advanced branch and bound algorithms used for NN verification, in particular, $\ABC$ \cite{ABC1,ABC2,betaCrown,chevalier2023gpuaccelerated}  for optimization-based bounds tightening (OBBT) 
to obtain tighter big-M bounds for the ReLU relaxations. Using better
big-M bounds tightens the MILP relaxation and improves the dual bound.

The second innovation concerns primal bounds: it aims at finding
strong primal solutions that can be used to warm-start the MILP-based
verifier. The proposed methodology uses gradient descent to quickly
find these primal solutions, which may dramatically improve the
convergence speed of the MILP solver. Indeed, the MILP-based verifiers
face challenges both in finding strong primal and dual bounds.  

The authors also implemented general cutting planes from
\cite{anderson2020strong} to further tighten the linear relaxation of
the MILP-based verifier.  These cuts were found ineffective, sometimes
even degrading performance on the larger grids.  These results are
thus omitted due to space considerations.


\subsection{$\ABC$  for Tighter Big-M Bounds}

\begin{figure}[!t]
    \centerline{\includegraphics[scale=.5]{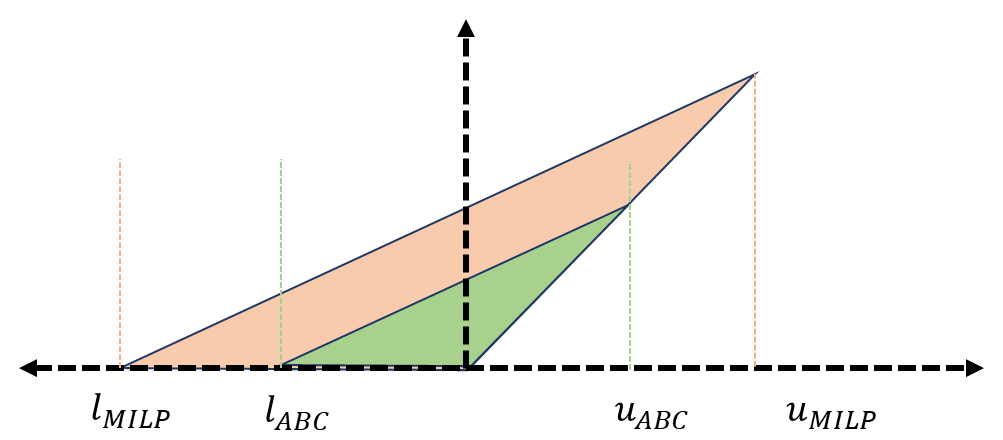}}
    \caption{Illustrates the feasible region for Relaxed ReLU. Here $y = ReLU(X)$. A tighter bound for the ReLU results in a tighter relaxation.}
    \label{fig_abc}
\end{figure}

The big-M reformulations of ReLU neurons, shown in equations
\eqref{Relu1}--\eqref{Relu5}, replace the ReLU activation function
with a mixed integer function bounded above and below.  For the MILP
algorithm to converge, the primal bound, which satisfies all the
constraints, including the binary variable constraints, must equal the
dual bound, a relaxed linear problem with some of the binary variables
defined as continuous variables $\in [0,1] $.  Tighter and more
accurate upper and lower bounds for the ReLU function result in a
tighter feasible region for the relaxed MILP problem, as shown in \cref{fig_abc}, thus resulting in a tighter dual bound.

Prior work on bound tightening for ReLU used interval bound propagation (IBP) and
MILP-based tightening. IBP bounds are commonly used to find the
upper and lower bounds of the ReLU function for NN verification: they
can be computed relatively quickly, but are usually loose, especially
as the depth of the network increases. MILP-based tightening
\cite{Andreas} has been proposed as an alternative to IBP for bound tightening.
MILP-based tightening must solve two
MILPs for finding the big-M bound at each node: one to find the lower
bound and one to find the upper bound.  The MILP problem for finding
the lower bound can be formulated as follows:
\begin{subequations}
\begin{align}
    \underline{\mathbf{Z}}_k  = \quad
    \min_{\pd} \quad
        & \hat{\mathbf{Z}}_k \\
    \text{s.t.} \quad
        & \eqref{NN1}, \eqref{Relu1}- \eqref{Relu5} \\
        & \pdmin \leq p^d \leq \pdmax
\end{align} 
\end{subequations}

\noindent
Similarly, the MILP problem for finding the upper bound can be formulated as follows: 
\begin{subequations}
\begin{align}
    \overline{\mathbf{Z}}_k  = \quad
    \max_{\pd} \quad
        & \hat{\mathbf{Z}}_k \\
    \text{s.t.} \quad
        & \eqref{NN1}, \eqref{Relu1}- \eqref{Relu5} \\
        & \pdmin \leq p^d \leq \pdmax
\end{align} 
\end{subequations}

\noindent
These MILPs must not be solved to optimality: instead, they may be run
for a few seconds, and the obtained bounds can be used as the big-M
values. Furthermore, this is a highly parallelizable task. However,
the MILP-based tightening is also a computationally intensive task for
large power systems.

As an alternative, this paper proposes the use of one of the top
branch and bound algorithms in the NN verification competitions
\cite{NNVerification}: $\ABC$ proposed in \cite{ABC1, betaCrown} to
achieve tighter big-M bounds. $\ABC$ is a fast and scalable NN
verifier that uses an efficient linear bound propagation, branch and
bound procedure, and GPU acceleration to find better bounds. Similar
to the MILP method proposed in \cite{Andreas}, $\ABC$ must be run
twice, once to get the upper bound, denoted by $u_{ABC}$, and the
lower bound, denoted by $l_{ABC}$, per ReLU. However, similar to the
MILP method for bound tightening, the verification can be parallelized
on GPUs.

\subsection{Gradient Descent for Strong Primal Solutions}

As mentioned earlier, on the primal side, general-purpose MIP solvers
also fail to quickly identify adversarial examples, i.e., input
values for which the NN prediction exhibits large constraint
violations. This hinders the overall solution process. To address this
issue, the paper introduces gradient-based attacks that leverage the
structure of the OPF problem and of the NN model.

The proposed methodology first identifies data points in the training
and test datasets that cause significant constraint violations.
These are used as starting points for a projected gradient ascent-based (PGA) local search,
wherein gradient steps are applied to iteratively modify the input values (i.e.,
the active power demand) to increase the constraint violation.
The use of multiple starting points improves the quality of the 
obtained adversarial examples, and makes use of the GPUs' parallelization capability.

The PGA algorithm is applied to the following optimization problem for the power balance violation
\begin{subequations}
\begin{align}
    \max_{\pd} \quad
        & \; \mathcal{L}^{\text{PB}} = v^{\text{PB}} \\
    \text{s.t.} \quad
        & v^{\text{PB}} = |\e^{\top} \pghat - e^{\top} \pd| \\
        & \eqref{NN1}, \eqref{Relu1}- \eqref{Relu5} \\
        & \pdmin \leq \pd \leq \pdmax
\end{align}
\end{subequations}

\noindent
Each iteration of the PGA algorithm can be formulated as follows:
\begin{subequations}
\begin{align}
    \pdadv_{i+1} &\gets \pdadv_{i} + \lambda \nabla_{\pd} \mathcal{L}^{\text{PB}}(\pdadv_{i})\\
    \pdadv_{i+1} &\gets \min( \max(\pdadv_{i+1}, \mathbf{\pdmin}),\mathbf{\pdmax})
\end{align}
\end{subequations}
where $\pdadv_{i+1}$ is the adversarial input after the
$i$th PGA iteration, and $\lambda$ is the
learning rate for the gradient ascent.
The initial value $\pdadv_{0}$ is the selected adversarial data point
taken from the training set.
Similarly, for the worst-case line flow violation, the optimization
problem can be formulated as follows:
\begin{subequations}
\begin{align}
        \max_{\pd} \quad
        & \; \mathcal{L}^{l} = v^{l} \\
    \text{s.t.} \quad
        & v^{l} = \max(0, |\pf| - \pfmax) \\
        & \pf = \Phi(\pghat - \pd)\\
        & \eqref{NN1}, \eqref{Relu1}- \eqref{Relu5} \\
        & \pdmin \leq \pd \leq \pdmax
\end{align}
\end{subequations}

Finally, the PGA algorithm can be implemented directly using backpropagation algorithms for NN training.
This allows the local heuristic to be executed on a GPU, enabling fast computations and parallel multi-starts.


%% file: tex/results.tex
\section{Numerical results}
\label{sec:results}

\newcommand{\ieee}{\texttt{ieee300}}
\newcommand{\goc}{\texttt{goc793}}
\newcommand{\pegaseS}{\texttt{pegase1k}}
\newcommand{\pegaseM}{\texttt{pegase2k}}

State-of-the-art techniques for extracting worst-case guarantees in
power system applications do not scale to large power systems and
complex neural networks. This section demonstrates how the proposed
acceleration schemes produce dramatic improvements compared to the
state-of-the-art in MILP-based NN verification. The results are
presented for DC-OPF problems and four different test power systems:
the \ieee, \goc, \pegaseS, and \pegaseM{} systems from the
PGLib-OPF network library v19.05 \cite{PGLib}.
The test system characteristics are given in Table \ref{TC}.
Namely, the table reports, for each system: the number of buses ($N_{b}$), loads ($N_{d}$), generators ($N_{g}$), and branches ($N_{l}$), and the maximum total active load in the dataset, in MW.

\begin{table}[!t]
\centering
\caption{The Characteristics of Test Cases.}
\label{TC}
\begin{tabular}{lrrrrr}
\toprule
Test Case 
    & \multicolumn{1}{c}{$N_b$}
    & \multicolumn{1}{c}{$N_d$}
    & \multicolumn{1}{c}{$N_g$}
    & \multicolumn{1}{c}{$N_l$}
    & $^\dagger$$\text{P}^{\text{max}}$ \\ 
\midrule
\ieee     & 300  
            & 199  
            & 57  
            & 411    
            & 23525                                                       
            \\
\goc     & 793  
            & 507  
            & 97  
            & 913   
            & 13198                                                  
            \\
\pegaseS    & 1354 
            & 674  
            & 260 
            & 1991   
            & 121766
            \\ 
\pegaseM    & 2869 
            & 1485 
            & 510 
            & 4582   
            & 132437                                                       
            \\ 
\bottomrule
\end{tabular}\\
{\footnotesize $^\dagger$Maximum total active load, in MW}
\end{table}

The training dataset is created by assuming that the demand at each
node varies between 60\% and 100\% of its nominal loading. The sum of
the maximum load over all nodes for each system is given in Table
\ref{TC}.
A datset of 10,000 random input values were generated for each
test case using Latin hypercube sampling \cite{hypercube}.  70\% of
them were assigned to the training dataset, 10\% to the validation
set, and 20\% to the unseen test dataset. MATPOWER \cite{MATPOWER} was
used to solve the DC-OPF problem and determine the optimal generation
set-points. The NN were implemented in PyTorch \cite{Pytorch} and used
the Adam optimizer \cite{Adam} for training, with a learning rate of
0.001. WandB \cite{wandb} was used to monitor and tune the
hyperparameters. The MILP-based NN verifiers were solved using the
Gurobi solver \cite{gurobi}. The NNs were trained on a
High-Performance Computing (HPC) server with an Intel Xeon E5-2650v4
processor and 256 GB RAM. The code and datasets to reproduce the
results are available online \cite{Code_git}.

\subsection{Average Neural Network Performance}

The experiments use NNs with three hidden layers with $N_{k}$ units to predict DC-OPF
solutions.
All NNs are trained until convergence and use early
stopping to prevent overfitting.
Table \ref{TabNNAvg} reports, for each system: the number of nodes in the hidden layers ($N_{k}$),
the average test loss ($\mathcal{L}^{\text{test}}_{0}$, in \%), the average power balance violations ($v^{\text{PB}}_{avg}$, in \%), and the average line flow violation ($v^{l}_{avg}$, in \%).
Recall that the loss function $\mathcal{L}$ is defined in Eq. \eqref{L_0}.
The relative constraint violations $v^{\text{PB}}_{avg}$ and $v^{l}_{avg}$ are expressed in percentage, relative to the maximum total load and each branch's line flow capacity, respectively.

\begin{table}[!t]
\centering
\caption{Average Neural Network Performance}
\label{TabNNAvg}
\begin{tabular}{lrrrr}
\toprule
Case 
    & \multicolumn{1}{c}{$N_{k}$}
    & \multicolumn{1}{c}{$\mathcal{L}^{\text{test}}_{0}$}
    & \multicolumn{1}{c}{$v^{\text{PB}}_{avg}$}
    & \multicolumn{1}{c}{$v^{l}_{avg}$}
 \\ 
\midrule
\ieee  & 50 
     & 0.29           
     & 0.23
     & 0.00 
\\ 
\goc  & 50         
     & 0.37             
     & 0.27
     & 0.00
\\ 
\pegaseS & 50   
     & 0.38           
     & 0.16 
     & 0.18
\\ 
\pegaseM & 100 
     & 0.42            
     & 0.21
     & 0.19
\\
\bottomrule
\end{tabular}
\end{table}

The results of Table \ref{TabNNAvg} show that all the NN models are performing adequately well, with less than 0.5 \% average prediction error in the test set.
Furthermore, all NN have low average constraint violations.
  
\subsection{The Impact of the Primal Acceleration}

Tables \ref{WC_PFB} and \ref{WC_LF} compare the primal bounds
available in the training dataset with those found by applying the
acceleration technique. The PGA algorithm was applied to
the fifty worst inputs in the dataset. The table reports the
warm-start primal bounds obtained in the dataset and with the
acceleration technique. It also reports the best primal and dual
bounds found by the MILP-based NN verifier in 5 hours.

The results show that the PGA acceleration identifies
data points with significantly larger worst-case violations than those
available in the dataset. In all cases but the \pegaseS{} problem, the
worst-case violations obtained using PGA were almost twice
as large as the worst-case data point from the dataset. Furthermore,
in the \ieee{} and \pegaseS{} problems, the data points generated by
PGA are close to the worst-case constraint violations. In
the \goc{} and \pegaseM{} problems, the MILP-based NN verification was unable to find significantly better solutions even after 5 hours. On
the \ieee{} and \pegaseS{} systems, the primal acceleration techniques
find near-optimal solutions. In the remaining test cases, it finds
primal solutions that are close the best-found primal solutions. The
primal acceleration is thus a critical tool for NN verification.
\begin{table}[!t]
\centering
\caption{Worst-Case Violations of the Power Flow Balance Constraints.}
\label{WC_PFB}
\begin{tabular}{lcccc}
    \toprule
    \multirow{2}{*}{Case} & \multicolumn{2}{c}{Warm-Start Primal Bounds} & \multicolumn{2}{c}{MILP} \\ 
    \cmidrule(lr){2-3} \cmidrule(lr){4-5}
                          & Dataset         & PGA        & Primal              & Dual             \\ 
    \midrule
    \ieee         & 6.24
                    & 13.64 
                    & 14.39
                    & 14.39                                                                      
    \\ 
    \goc             & 8.68 
                    & 28.89  
                    & 29.71  
                    & 61.76
    \\
    \pegaseS
        & 24.48 
        & 24.67 
        & 24.69 
        & 24.69
    \\
    \pegaseM
        & 11.91
        & 23.00
        & 23.01
        & 77.75
    \\
    \bottomrule
\end{tabular}\\
{\footnotesize All values are in \% w.r.t the max loading.}
\end{table}

\begin{table}[!t]
\centering
\caption{Worst-Case Violations of the Line Flow Constraints.}
\label{WC_LF}
\begin{tabular}{lccrr}
    \toprule
    & \multicolumn{2}{c}{Warm-Start Primal Bounds} 
    & \multicolumn{2}{c}{MILP}
\\
    \cmidrule(lr){2-3} \cmidrule(lr){4-5}
    Case 
        & \multicolumn{1}{c}{Dataset}
        & \multicolumn{1}{c}{PGA}    
        & \multicolumn{1}{c}{Primal}              
        & \multicolumn{1}{c}{Dual}             \\ 
    \midrule
\ieee     &  \phantom{0}0.00 
            &  \phantom{0}0.00
            &  \phantom{0}0.00  
            &  \phantom{0}0.00                                                                      \\ 
\goc     & 20.87  
            & 35.34       
            & 36.16 
            &  64.27                                                                          \\ 
\pegaseS    & \phantom{0}1.50
            & \phantom{0}1.53 
            & \phantom{0}1.54    
            & \phantom{0}1.54                                                                          \\ 
\pegaseM    & 17.02 
            & 41.64
            & 42.45 
            & 123.54                                                                           \\ \hline
\end{tabular}\\
    {\footnotesize All values are in \% w.r.t the line capacity.}
\end{table}
\subsection{The Impact of the Dual Acceleration}

This section compares the best dual bound that the Gurobi solver was
able to obtain after solving for 5 hours when using different
optimization-based bounds tightening (OBBT) techniques: Interval Bound Propagation
(IBP), the MILP formulation from \cite{Andreas}, and $\ABC$ for
getting the RELU big-M bounds. Both $\ABC$ and the MILP algorithm for
finding big-M bounds were run for 10s per ReLU. This is a highly
parallelizable task that can be run on multiple GPUs in the case of
$\ABC$ and CPUs in the case of MILPs.

{\em Remarkably, the results show that the proposed methodology
  converges on the \ieee{} and \pegaseS{} systems. This was impossible
  with the state-of-the-art OBBT techniques like MILP-based OBBT or
  IBP.} Furthermore, in \pegaseM{} and \goc, the big-M bound given by
$\ABC$ helped significantly close the gap between the primal and dual
bounds. This shows the substantial value of $\ABC$ for NN
verification.  problem. Similarly, for worst-case line flow constraint
violations, the proposed approach converges on the \pegaseS{} system,
which is out of reach for the state-of-the-art. It also closes the gap
between the primal and dual bounds on the \pegaseM{} and \goc {}
systems. When comparing the computational times for the cases that
converged in 5 hours, Table \ref{ComTime} shows that $\ABC$ was able
to converge much faster than the other technique.

\begin{table}[!t]
    \centering
    \caption{Worst Case Power Flow Balance}
    \label{WC_Dual_PFB}
    \begin{tabular}{lcccc}
        \toprule
            & \multicolumn{3}{c}{OBBT Technique} & \multicolumn{1}{c}{MILP}\\
        \cmidrule(lr){2-4} \cmidrule(lr){5-5}
        Case
            & IBP 
            & MILP 
            & $\ABC$ 
            & Primal \\
        \midrule
        \ieee
            & 41.33
            & 31.25
            & 14.39
            & 14.39
        \\
        \goc
            & 86.08
            & 70.70
            & 61.76
            & 29.71
        \\
        \pegaseS
            & 48.27
            & 37.68
            & 24.69
            & 24.69
            \\
        \pegaseM
            & 93.49
            & 86.81
            & 77.75
            & 23.01
            \\
        \bottomrule
    \end{tabular}\\
    {\footnotesize All values are in \% w.r.t the max loading.}
\end{table}

\begin{table}[!t]
    \centering
    \caption{Worst Case Line Flow Constraint Violation}
    \label{WC_Dual_LFV}
    \begin{tabular}{lrrrr}
        \toprule
            & \multicolumn{3}{c}{OBBT Technique} & \multicolumn{1}{c}{MILP}\\
        \cmidrule(lr){2-4} \cmidrule(lr){5-5}
        Case
            & IBP 
            & MILP 
            & $\ABC$ 
            & Primal \\
        \midrule
        \ieee
            & 0.00
            & 0.00
            & 0.00
            & 0.00
        \\
        \goc
            & 137.42
            & 65.94
            & 64.27
            & 36.16                                                                         \\
        \pegaseS
            & 9.27
            & 4.72
            & 1.54
            & 1.54
            \\
        \pegaseM
            & 162.96
            & 130.85
            & 123.54
            & 42.45
            \\
        \bottomrule
    \end{tabular}\\
    {\footnotesize All values are in \% w.r.t the line capacity.}
\end{table}

\begin{table}[!t]
\centering
\caption{Computing time statistics}
\label{ComTime}
\begin{tabular}{lcrrr}
\toprule
     &
     & \multicolumn{3}{c}{OBBT technique} \\
\cmidrule{3-5} 
    Case &          
        & \multicolumn{1}{c}{IBP}
        & \multicolumn{1}{c}{MILP}
        & \multicolumn{1}{c}{$\ABC$}
\\
\midrule
\ieee 
    & $v_{PB}$ & -          & -         & 7595.00      \\                   
    & $v_{l}$  & 5.79       & 3.27      & 1.59      \\ 
\midrule
\pegaseS & $v_{PB}$ & -          & -         & 147.00       \\ 
                      & $v_{l}$  & 7244.02    & 7234.50    & 2114.46   \\
\bottomrule
\end{tabular}\\
    {\footnotesize All values are in seconds. In case of $v_{PB}$ both for \ieee{} and \pegaseS{} system only the MILP-based verification that used $\ABC$ for OBBT managed to converge.}
\end{table}

%% file: pscc2024_paper.bbl
\begin{thebibliography}{10}
\providecommand{\url}[1]{#1}
\csname url@samestyle\endcsname
\providecommand{\newblock}{\relax}
\providecommand{\bibinfo}[2]{#2}
\providecommand{\BIBentrySTDinterwordspacing}{\spaceskip=0pt\relax}
\providecommand{\BIBentryALTinterwordstretchfactor}{4}
\providecommand{\BIBentryALTinterwordspacing}{\spaceskip=\fontdimen2\font plus
\BIBentryALTinterwordstretchfactor\fontdimen3\font minus
  \fontdimen4\font\relax}
\providecommand{\BIBforeignlanguage}[2]{{%
\expandafter\ifx\csname l@#1\endcsname\relax
\typeout{** WARNING: IEEEtran.bst: No hyphenation pattern has been}%
\typeout{** loaded for the language `#1'. Using the pattern for}%
\typeout{** the default language instead.}%
\else
\language=\csname l@#1\endcsname
\fi
#2}}
\providecommand{\BIBdecl}{\relax}
\BIBdecl

\bibitem{CARPENTIER1979}
J.~Carpentier, ``Optimal power flows,'' \emph{International Journal of
  Electrical Power \& Energy Systems}, vol.~1, no.~1, pp. 3--15, 1979.

\bibitem{MLDCOPF}
D.~Deka and S.~Misra, ``Learning for dc-opf: Classifying active sets using
  neural nets,'' in \emph{2019 IEEE Milan PowerTech}, 2019, pp. 1--6.

\bibitem{ReviewMLSec}
O.~A. Alimi, K.~Ouahada, and A.~M. Abu-Mahfouz, ``A review of machine learning
  approaches to power system security and stability,'' \emph{IEEE Access},
  vol.~8, pp. 113\,512--113\,531, 2020.

\bibitem{SurveyML}
F.~Hasan, A.~Kargarian, and A.~Mohammadi, ``A survey on applications of machine
  learning for optimal power flow,'' in \emph{2020 IEEE Texas Power and Energy
  Conference (TPEC)}, 2020, pp. 1--6.

\bibitem{recent}
L.~Duchesne, E.~Karangelos, and L.~Wehenkel, ``Recent developments in machine
  learning for energy systems reliability management,'' \emph{Proceedings of
  the IEEE}, vol. 108, no.~9, pp. 1656--1676, 2020.

\bibitem{surrogate}
A.~Kody, S.~Chevalier, S.~Chatzivasileiadis, and D.~Molzahn, ``Modeling the ac
  power flow equations with optimally compact neural networks: Application to
  unit commitment,'' \emph{Electric Power Systems Research}, vol. 213, p.
  108282, 2022.

\bibitem{Andreas}
A.~Venzke, G.~Qu, S.~Low, and S.~Chatzivasileiadis, ``Learning optimal power
  flow: Worst-case guarantees for neural networks,'' in \emph{2020 IEEE
  International Conference on Communications, Control, and Computing
  Technologies for Smart Grids (SmartGridComm)}.\hskip 1em plus 0.5em minus
  0.4em\relax IEEE, 2020, pp. 1--7.

\bibitem{MYDC}
R.~Nellikkath and S.~Chatzivasileiadis, ``{Physics-Informed Neural Networks for
  Minimising Worst-Case Violations in} {DC} {Optimal} {Power} {Flow},'' in
  \emph{2021 IEEE International Conference on Communications, Control, and
  Computing Technologies for Smart Grids (SmartGridComm)}, 2021, pp. 419--424.

\bibitem{MYAC}
\BIBentryALTinterwordspacing
------, ``{Physics-Informed Neural Networks for AC Optimal Power Flow},''
  \emph{Electric Power Systems Research}, vol. 212, p. 108412, 2022. [Online].
  Available:
  \url{https://www.sciencedirect.com/science/article/pii/S0378779622005636}
\BIBentrySTDinterwordspacing

\bibitem{sam_stt}
S.~Chevalier and S.~Chatzivasileiadis, ``Global performance guarantees for
  neural network models of ac power flow,'' \emph{arXiv preprint
  arXiv:2211.07125}, 2022.

\bibitem{MinWC}
R.~Nellikkath and S.~Chatzivasileiadis, ``Minimizing worst-case violations of
  neural networks,'' \emph{arXiv preprint arXiv:2212.10930}, 2022.

\bibitem{Enrich}
------, ``Enriching neural network training dataset to improve worst-case
  performance guarantees,'' \emph{arXiv preprint arXiv:2303.13228}, 2023.

\bibitem{ABC1}
\BIBentryALTinterwordspacing
K.~Xu, H.~Zhang, S.~Wang, Y.~Wang, S.~Jana, X.~Lin, and C.-J. Hsieh, ``{Fast
  and Complete}: Enabling complete neural network verification with rapid and
  massively parallel incomplete verifiers,'' in \emph{International Conference
  on Learning Representations}, 2021. [Online]. Available:
  \url{https://openreview.net/forum?id=nVZtXBI6LNn}
\BIBentrySTDinterwordspacing

\bibitem{ABC2}
\BIBentryALTinterwordspacing
H.~Zhang, T.-W. Weng, P.-Y. Chen, C.-J. Hsieh, and L.~Daniel, ``Efficient
  neural network robustness certification with general activation functions,''
  \emph{Advances in Neural Information Processing Systems}, vol.~31, pp.
  4939--4948, 2018. [Online]. Available:
  \url{https://arxiv.org/pdf/1811.00866.pdf}
\BIBentrySTDinterwordspacing

\bibitem{betaCrown}
S.~Wang, H.~Zhang, K.~Xu, X.~Lin, S.~Jana, C.-J. Hsieh, and J.~Z. Kolter,
  ``{Beta-CROWN}: Efficient bound propagation with per-neuron split constraints
  for complete and incomplete neural network verification,'' \emph{Advances in
  Neural Information Processing Systems}, vol.~34, 2021.

\bibitem{chevalier2023gpuaccelerated}
S.~Chevalier, I.~Murzakhanov, and S.~Chatzivasileiadis, ``Gpu-accelerated
  verification of machine learning models for power systems,'' 2023.

\bibitem{DCOPF}
B.~Stott, J.~Jardim, and O.~Alsac, ``Dc power flow revisited,'' \emph{IEEE
  Transactions on Power Systems}, vol.~24, no.~3, pp. 1290--1300, 2009.

\bibitem{anderson2020strong}
R.~Anderson, J.~Huchette, W.~Ma, C.~Tjandraatmadja, and J.~P. Vielma, ``Strong
  mixed-integer programming formulations for trained neural networks,''
  \emph{Mathematical Programming}, vol. 183, no. 1-2, pp. 3--39, 2020.

\bibitem{NNVerification}
M.~N. M{\"u}ller, C.~Brix, S.~Bak, C.~Liu, and T.~T. Johnson, ``The third
  international verification of neural networks competition (vnn-comp 2022):
  summary and results,'' \emph{arXiv preprint arXiv:2212.10376}, 2022.

\bibitem{PGLib}
S.~Babaeinejadsarookolaee, A.~Birchfield, R.~D. Christie, C.~Coffrin,
  C.~DeMarco, R.~Diao, M.~Ferris, S.~Fliscounakis, S.~Greene, R.~Huang
  \emph{et~al.}, ``The power grid library for benchmarking ac optimal power
  flow algorithms,'' \emph{arXiv preprint arXiv:1908.02788}, 2019.

\bibitem{hypercube}
M.~D. McKay, R.~J. Beckman, and W.~J. Conover, ``A comparison of three methods
  for selecting values of input variables in the analysis of output from a
  computer code,'' \emph{Technometrics}, vol.~42, no.~1, pp. 55--61, 2000.

\bibitem{MATPOWER}
R.~D. Zimmerman, C.~E. Murillo-S{\'a}nchez, and R.~J. Thomas, ``Matpower:
  Steady-state operations, planning, and analysis tools for power systems
  research and education,'' \emph{IEEE Transactions on power systems}, vol.~26,
  no.~1, pp. 12--19, 2010.

\bibitem{Pytorch}
\BIBentryALTinterwordspacing
A.~Paszke, S.~Gross, F.~Massa, A.~Lerer, J.~Bradbury, G.~Chanan, T.~Killeen,
  Z.~Lin, N.~Gimelshein, L.~Antiga, A.~Desmaison, A.~Kopf, E.~Yang, Z.~DeVito,
  M.~Raison, A.~Tejani, S.~Chilamkurthy, B.~Steiner, L.~Fang, J.~Bai, and
  S.~Chintala, ``Pytorch: An imperative style, high-performance deep learning
  library,'' in \emph{Advances in Neural Information Processing Systems 32},
  H.~Wallach, H.~Larochelle, A.~Beygelzimer, F.~d\textquotesingle
  Alch\'{e}-Buc, E.~Fox, and R.~Garnett, Eds.\hskip 1em plus 0.5em minus
  0.4em\relax Curran Associates, Inc., 2019, pp. 8024--8035. [Online].
  Available:
  \url{http://papers.neurips.cc/paper/9015-pytorch-an-imperative-style-high-performance-deep-learning-library.pdf}
\BIBentrySTDinterwordspacing

\bibitem{Adam}
\BIBentryALTinterwordspacing
D.~P. Kingma and J.~Ba, ``Adam: {A} method for stochastic optimization,'' in
  \emph{3rd International Conference on Learning Representations, {ICLR} 2015,
  San Diego, CA, USA, May 7-9, 2015, Conference Track Proceedings}, Y.~Bengio
  and Y.~LeCun, Eds., 2015. [Online]. Available:
  \url{http://arxiv.org/abs/1412.6980}
\BIBentrySTDinterwordspacing

\bibitem{wandb}
\BIBentryALTinterwordspacing
L.~Biewald, ``Experiment tracking with weights and biases,'' 2020, software
  available from wandb.com. [Online]. Available: \url{https://www.wandb.com/}
\BIBentrySTDinterwordspacing

\bibitem{gurobi}
\BIBentryALTinterwordspacing
{Gurobi Optimization, LLC}, ``{Gurobi Optimizer Reference Manual},'' 2022.
  [Online]. Available: \url{https://www.gurobi.com}
\BIBentrySTDinterwordspacing

\bibitem{Code_git}
\BIBentryALTinterwordspacing
P.~V.~H. R.~Nellikkath, M.~Tanneau and S.~Chatzivasileiadis, ``Supplementary
  data and code: Scalable exact verification of optimization proxies for
  large-scale optimal power flow,'' 2023. [Online]. Available:
  \url{https://github.com/RahulNellikkath/ScalableExactVerification}
\BIBentrySTDinterwordspacing

\end{thebibliography}
